\documentclass[twocolumn,10pt,cleanhead,cleanfoot]{asme2e}

\usepackage[usenames, dvipsnames]{color}
\usepackage{graphicx}
\usepackage{siunitx}
\usepackage{cite}
\usepackage{subfigure}
\usepackage{array}
\usepackage{afterpage}
\usepackage{amsfonts}
\usepackage{mathtools}
\usepackage{amsmath}
\usepackage{amssymb}
\usepackage{algpseudocode}
\usepackage{algorithm}
\usepackage{relsize}
\usepackage{mathrsfs} 
\usepackage{todonotes}
\usepackage{color}
\usepackage{ulem} 						%
\usepackage{xspace}

\usepackage{parskip}
\usepackage[T1]{fontenc}

\confdate{26-29}
\confmonth{August}
\confyear{2018}
\confcity{Quebec}
\confcountry{Canada}

\papernum{DETC2018-85594}

\title{Draft: On Locomotion of a Laminated Fish-inspired robot in a Small-to-size Environment}

\author{Mohammad Sharifzadeh, Roozbeh Khodambashi, Wenlong Zhang, Daniel Aukes
    \affiliation{
	\small Polytechnic School, Ira A. Fulton Schools of Engineering\\
	\small Arizona State University\\
	\small Mesa, Arizona, 85212\\
    \small Email: \{sharifzadeh,rkhodamb,wenlong.zhang,daukes\}@asu.edu
    }	
}

\begin{document}

\maketitle    

\begin{abstract}
{\it Many different robots have been designed and built to work under water. In many cases, researchers have chosen to use bio-inspired platforms. In most cases, the main goal of the fish inspired robots has been set to autonomously swim and maneuver in an environment spacious compared to the fish's size. In this paper, the identification \& control of a low-cost fish-inspired robot is studied with goal of building a mechanism to not only swim in water but able to interact with its narrow environment. 
The robotic fish under study uses tail propulsion as main locomotion. Moreover, proper propulsion regimes are identified and used to model and control thrust generated by propulsion.
}
%
%

\end{abstract}




\section{INTRODUCTION}
Autonomous Underwater Vehicles (AUVs) are widely studied for operating below the surface of the water.  
Underwater propulsion is one of the main components of AUVs and underwater robots. This type of propulsion is a hard problem to understand analytically due to the highly nonlinear and turbulent nature of the water currents around an object. These turbulent flows result in the loss of energy and therefore energy efficient propulsion systems are of special interest for many researchers. 

Evolution provides a variety of successful underwater locomotion examples. Different fish species have developed different mechanisms of underwater propulsion which are energy efficient; many species use fins to maneuver through the water. Therefore, it is not surprising that the fish interest researchers for their efficiency and their maneuverability~\cite{r20}. A fish can make half-turn on 1/10 its length without losing speed, whereas a submarine needs 10 times its length by slowing down half to do the same. 
 The research on robotic fish are mainly focused on dynamic modeling~\cite{r6,r7,r8,r9} and control~\cite{r10,r11,r12,r13,r14,r15}. An extensive review and classification of different fish species and marine robots inspired by them is given in~\cite{r5}. Several other researchers have studied the effect of a certain parameter on the performance of a  robot. For example, \cite{r16} has addressed the effect of artificial caudal fins on the fish robot's performance. Marchese et al \& Zhong et al have used soft materials plus soft robotic fabrication techniques to manufacture a robot fish~\cite{r17,r18}.  In~\cite{r19}, Liu et al study the movement of a carangiform fish and then a robotic fish is controlled based on the obtained results. Turning performance of a robotic fish inspired by a sea bream is studied in~\cite{r20}. A parametric study of a fish robot performance is carried on in~\cite{param}. This paper used a 6-meter long water tunnel with flow control over the test tank to provide a known current for robotic fish understudy. Moreover, they used a robot with the tail consisting of one active joint and two passive joints as well as a flexible caudal fin. Similar to the previous research, in most cases, conventional robotics is unable to replicate specific features of fish locomotion, such as manufacturing a smooth and continuously flexible hull~\cite{r2}.
 This issue can be solved using recently-developed laminate manufacturing techniques~\cite{r3,r4}. 
 Throughout all these papers, the authors have identified an opportunity to address fish-inspired locomotion in the tail using laminate techniques, which have the capability of being able to be rapidly prototyped and tested. In addition, the trend in fish inspired robot is to eliminate the systematic design approach using soft materials or to utilize traditional robotic fabrication techniques with a limited number of passive joints. Using laminate techniques, it is possible to easily add the desired number of passive joints in desired positions, as well as using different materials and thicknesses to achieve desired stiffness. 

\begin{figure}[t!]
\centering
\includegraphics[scale = 0.135]{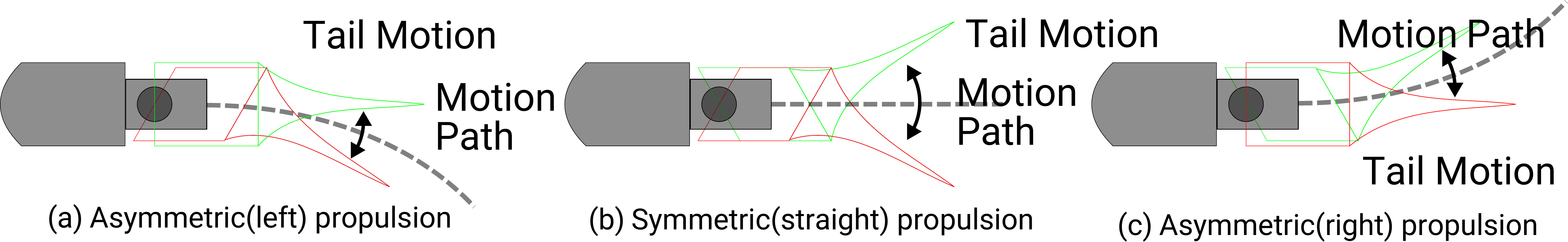}
\caption{Maneuvering strategy and effect of tail movement on generated force}
\label{pic_Strat}
\end{figure}

The main contribution of this study is to investigate the locomotion of a fish-inspired robot swimming in a tight workspace
and interacting with environment, while using forward propulsion in a high-flow channel to apply force to an object. 
This work is motivated by the need for a small, low-power robotic platform which can navigate autonomously in small canals for cleaning, maintenance, and inspection tasks.
 The focus of this paper is to use a tail fin manufactured via laminate fabrication techniques (known as Caudal fin in fish morphology) for locomotion. 
 Laminate fabrication techniques permits iterating rapidly through a complex design space to tune the stiffness \& damping properties of soft flexure hinges between a variable number of rigid segments, in order to quickly fabricate and validate an optimal fin design. Laminate fabrication methods are also low-cost, with the structure and tail components (\$5) costing a fraction of the selected actuator (\$55). 
Even though the laminate platform is novel, the focus of this paper is 
to facilitate a deeper understanding of the control issues at play in small environments and narrow passages.

The paper is organized as follows: Section 2 will describe maneuvering strategies as well as test setup construction. In Section 3, design and manufacturing of the robotic fish hardware is discussed. A model of the generated force  by robotic fish propulsion has been developed which is presented in section 4. In Section 5, different controllers are designed based on the identified model of the robotic fish. The paper concludes with some remarks and suggestions for future work indicated by obtained results.


\section{Maneuvering Strategies \& Experimental Setup}
\begin{table}[t]
\caption{List of Propulsion parameters}
\label{table_var}
\begin{center}
\begin{tabular}{c| c}
Symbol & Definition \\
\hline
$a$ & Propulsion Amplitude\\
\hline
$b$ & Offset in asymmetric propulsion\\
\hline
$f$ & Propulsion frequency\\
\hline
$t$ & Propulsion time\\
\hline
$\theta$ & Angle of servo actuator in propulsion\\
\hline
\end{tabular}
\end{center}
\end{table}
In nature, swimming modes of fish can be divided into two main categories, namely caudal fin locomotion and pectoral fin locomotion~\cite{fish_general}. In this paper, propulsion caused by caudal fin is considered as the main focus of our locomotion strategy. Figure~\ref{pic_Strat} shows the locomotion strategy proposed in this paper.
The fin actuator is commanded to follow the following angle: 
\begin{equation}
\theta(t) = b + a \sin(2\pi f t)
\label{eq_servo}
\end{equation}
Description of the propulsion parameters are described in Table~\ref{table_var}
The design goal is to enable the fish to interact with its environment, including vegetation, which may be present in shallow, narrow waterways of the southwestern United States. The interaction can involve pushing an object or cutting a specimen from a plant. The authors believe that this working environment, which contains stationary propulsion in a narrow canal, is challenging due to reflected waves coming back from canal walls, which can impact sensing navigation by imparting disturbances on the robot caused either by the world or the robot itself.
\begin{figure}[t]
\centering
\includegraphics[scale = 0.3]{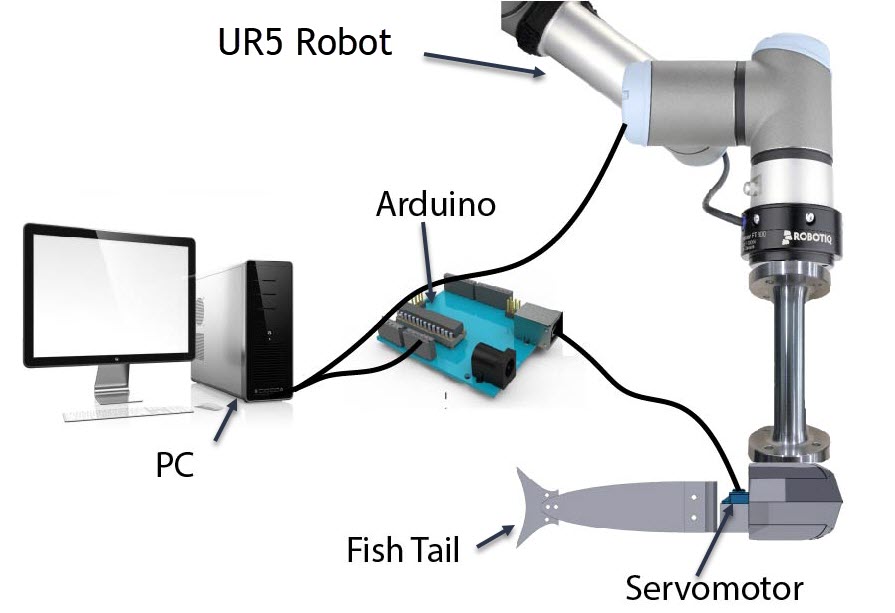}
\caption{Test setup for identification \& control}
\label{pic_setup}
\end{figure}

A test setup has been designed for evaluating the performance of underwater, robotic fish designs by monitoring the thrust generated from undulatory swimming. The design of the test setup is based on the overall strategy for controlling the robot and its working environment. 
Figure~\ref{pic_setup} shows the experimental setup. The robotic fish is attached to an force-torque sensor (ROBOTIQ FT300) using an aluminum bar. The torque applied by the robotic fish on the sensor through the attachment bar is measured and converted to force by dividing it by the length of the attachment bar (distance of the application point of the force to the origins of the sensor), since the force measurement alone is small relative to the range of the sensor. A 0.3048\si{\metre} L $\times$ 0.1524\si{\metre} W $\times$ 0.254\si{\metre} H water tank is used to model a small canal work environment.
The robot is attached to a UR5 robotic arm which can move the fish in the water with constant speed to simulate water current.

An Arduino is responsible for controlling the servomotor. Force and torque data is transferred to the computer through the Modbus protocol. A Python script is responsible for sending motor commands to the Arduino and reading data from the force and torque sensor.\footnote{Details of the Python code are presented in Appendix A.}

\begin{figure}[t]
\centering
\includegraphics[scale = 0.37]{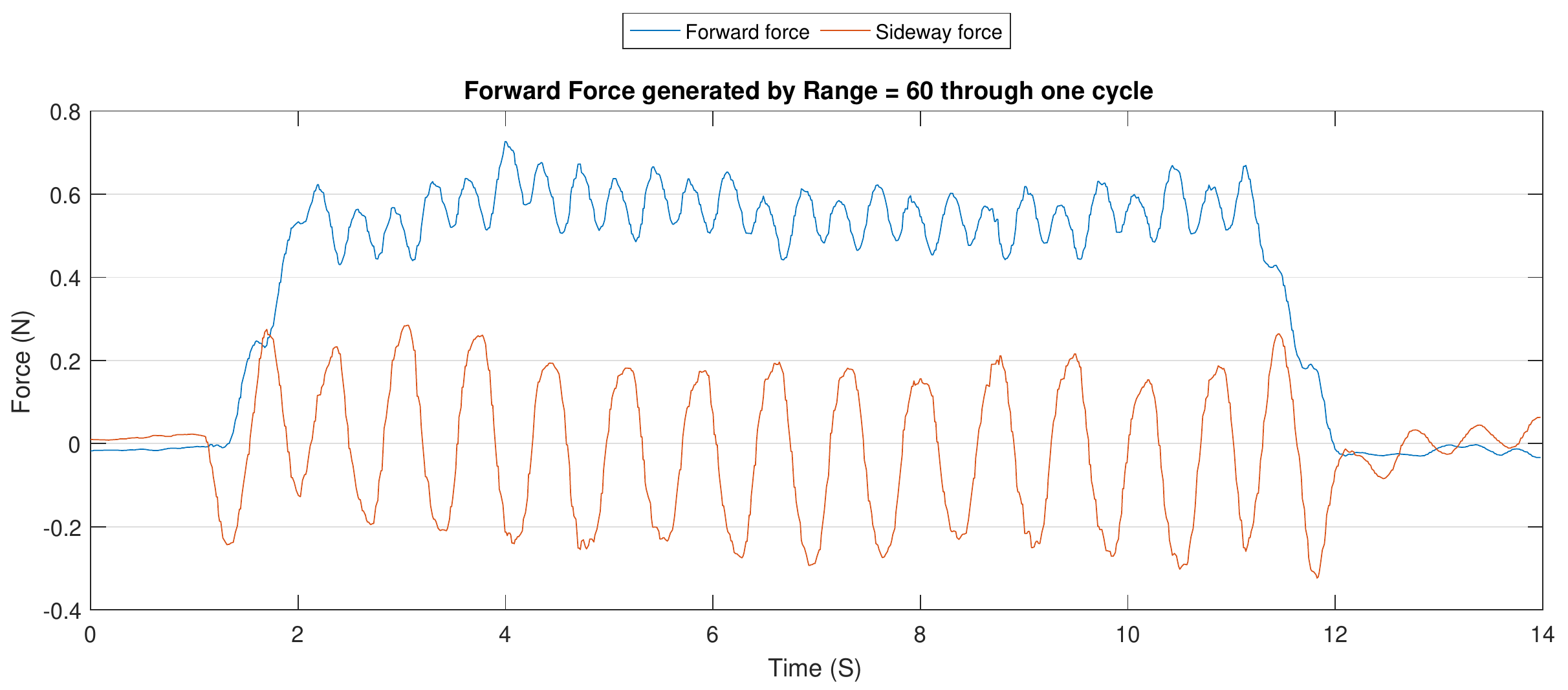}
\caption{Forward and lateral force generated by symmetric propulsion.}
\label{pic_sample}
\end{figure}

Figure~\ref{pic_sample} shows 10 seconds of raw forward and side-force data for a symmetric undulatory motion with amplitude of 60deg, in order to provide the reader with insight on the forces generated via fin propulsion.
 In this study, the forces generated by the fin are split into two main elements corresponding to the fish forward and sideway (lateral) directions. According to this figure, the generated forces are periodic in nature, as was expected due to the undulatory motion of the tail. It can be observed that although the sideway force is changing, the average force  produced over a cycle is close to zero due to fins symmetric propulsion. The average forward force is also positive although there are periodic variations observed in it. As a result,  force-torque data is averaged over a window of specified number of cycles. In this method, the average amount of forward and lateral forces generated by the tail are stored for subsequent use. 
%
%

\section{Design \& Prototyping}
\begin{figure}[t]
\centering
\subfigure[Propulsion generated force measurement prototype.]{\includegraphics[scale=0.35]{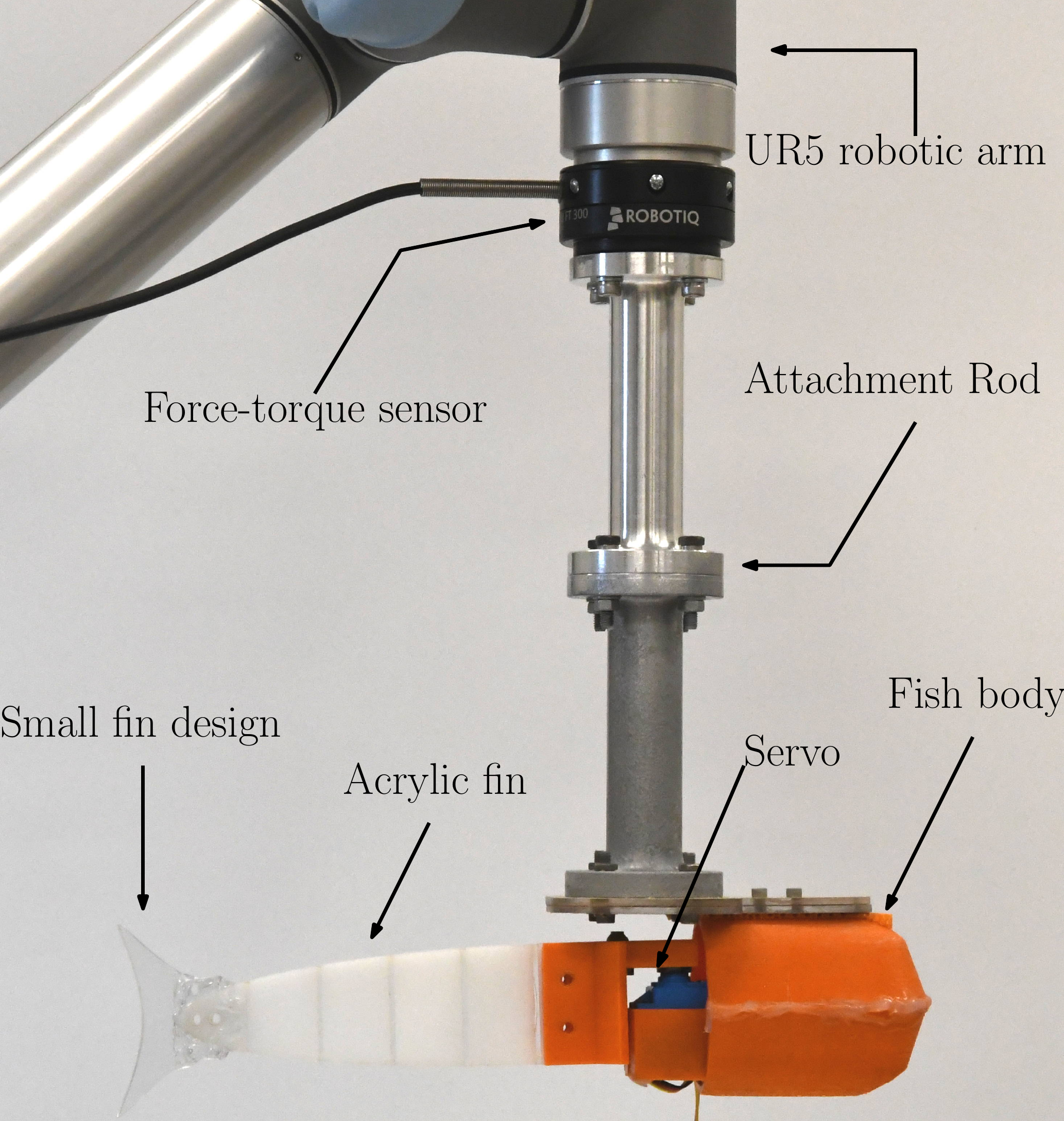}
\label{pic_fixed}
}
\subfigure[The float attached to the fish prototype to help in floating it in the water.]{\includegraphics[scale = 0.35]{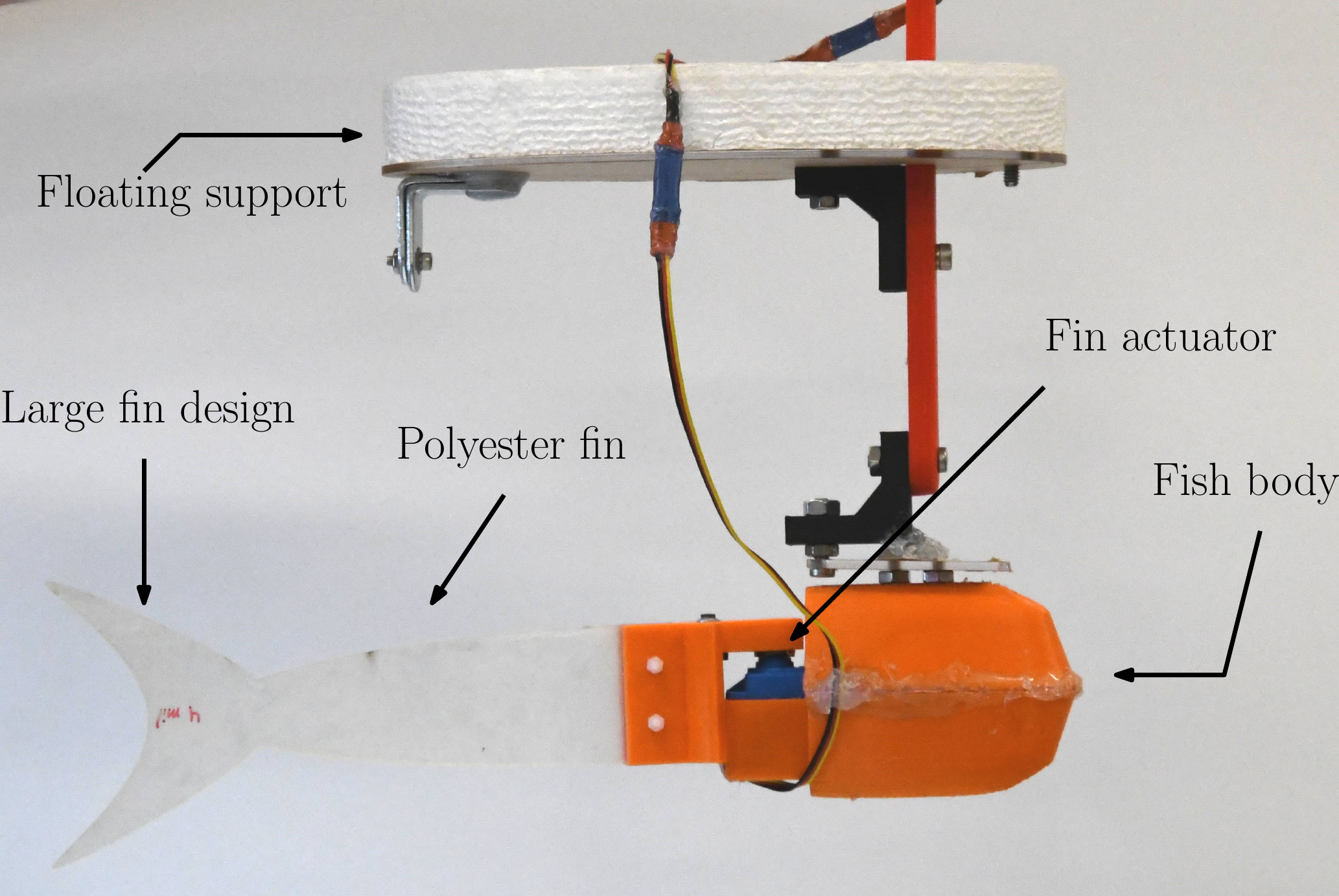}
\label{pic_moving}
}
\caption{Prototypes made to evaluate generating force and swim ability.}
\end{figure}

In designing the caudal fin, the results of the parametric design study conducted in~\cite{param} were considered, specifically the aspect ratio of the end fin. Initially, a laminated design was proposed with one actuated joint and five passive joints to provide the undulatory movement of the fin in the water. The prototype based on this design was built with acrylic (Fig.~\ref{pic_fixed}). Initial testing revealed that the thrust generated by the fin was not as high as hoped. The authors believe that the main reason was due to the fact that the tail was heavy and also thick. As a result, for further prototypes, the whole fin was built from sheets of polyester laminated together in different numbers to provide a variety of stiffnesses for evaluation.  Individual sheets were laminated together and then cut in a CO$_2$ laser.

The main components of the robotic fish locomotor are the tail and the actuator which are shown in Fig.~\ref{pic_fixed}.
A waterproof servomotor was selected based on the torque and speed required to provide thrust via the tail. The motor holder and the body of the fish are made using 3D printed parts. A floating support was made using foam to help the fish to float in water (Fig.~\ref{pic_moving}).


Table~\ref{table_Material} shows the average amount of force generated in all directions for different fin materials and designs. The result shows an increase in generated force with stiffer material while confirming the assertion made in~\cite{param} about producing more force with Larger back fin~(Fig.~\ref{pic_moving}). Moreover, the fins were installed and the robot's ability to swim was tested based on observation. 
One key observation was that as fin stiffness went up, rotation in the tail led to more rotation in the main body of the fish, rendering higher tail forces less useful at generating forward thrust.  Thus, a final material thickness of 1.016 mm was used for subsequent fin designs.
\begin{table}
\centering
\caption{Different design \& material effect.}
\label{table_Material}
 \begin{tabular}{>{\centering\arraybackslash}m{2.5cm} >{\centering\arraybackslash}m{1.5cm} >{\centering\arraybackslash}m{2.5cm} >{\centering\arraybackslash}m{1.0cm}}
\hline
Material \& width (mm) 
& 
Fin Design
&
Mean generated force
&
Swim ability
\\ \hline
6.35 Acrylic
&
Small
&
0.1983
&
Yes
\\ \hline
0.254 polyester
&
Small
&
0.2361
&
Yes
\\ \hline
0.254 polyester
&
Large
&
0.2647
&
Yes
\\ \hline
0.508 polyester
&
Small
&
0.2944
&
Yes
\\ \hline
0.508 polyester 
&
Large
&
0.3154
&
Yes
\\ \hline
1.016 polyester
&
Large
&
0.551
&
Yes
\\ \hline
1.524 polyester 
&
Large
&
1.157
&
No
\\ \hline
\end{tabular}
\end{table}

\section{Identification \& Model Extraction}

The goal of this section is to discuss the identification of  the magnitude and direction of thrust which can be generated via changing control signals in the robotic fish platform. To this end, providing a model that relates the input and output of the system is studied. 
According to~\cite{param}, flapping amplitude and frequency were the two control parameters found most effective at generating forward motion and thrust. 
Based on our design, placing the servo at the base of the tail, this permits us to select the servo's angle as an input variable. 
The input to the system is selected to be the amplitude and range of the undulatory motion of the servomotor. Moreover, angular offset from the symmetric plane has the most effect on direction of the generated force (Fig~\ref{pic_Strat}). As mentioned before, the output of the system is the force generated by the tail propulsion, which is measured by the F/T sensor. 

\begin{figure*}[t]
\centering
\subfigure[Generated forward force as a function of range, offset and frequency]{\includegraphics[scale=0.6]{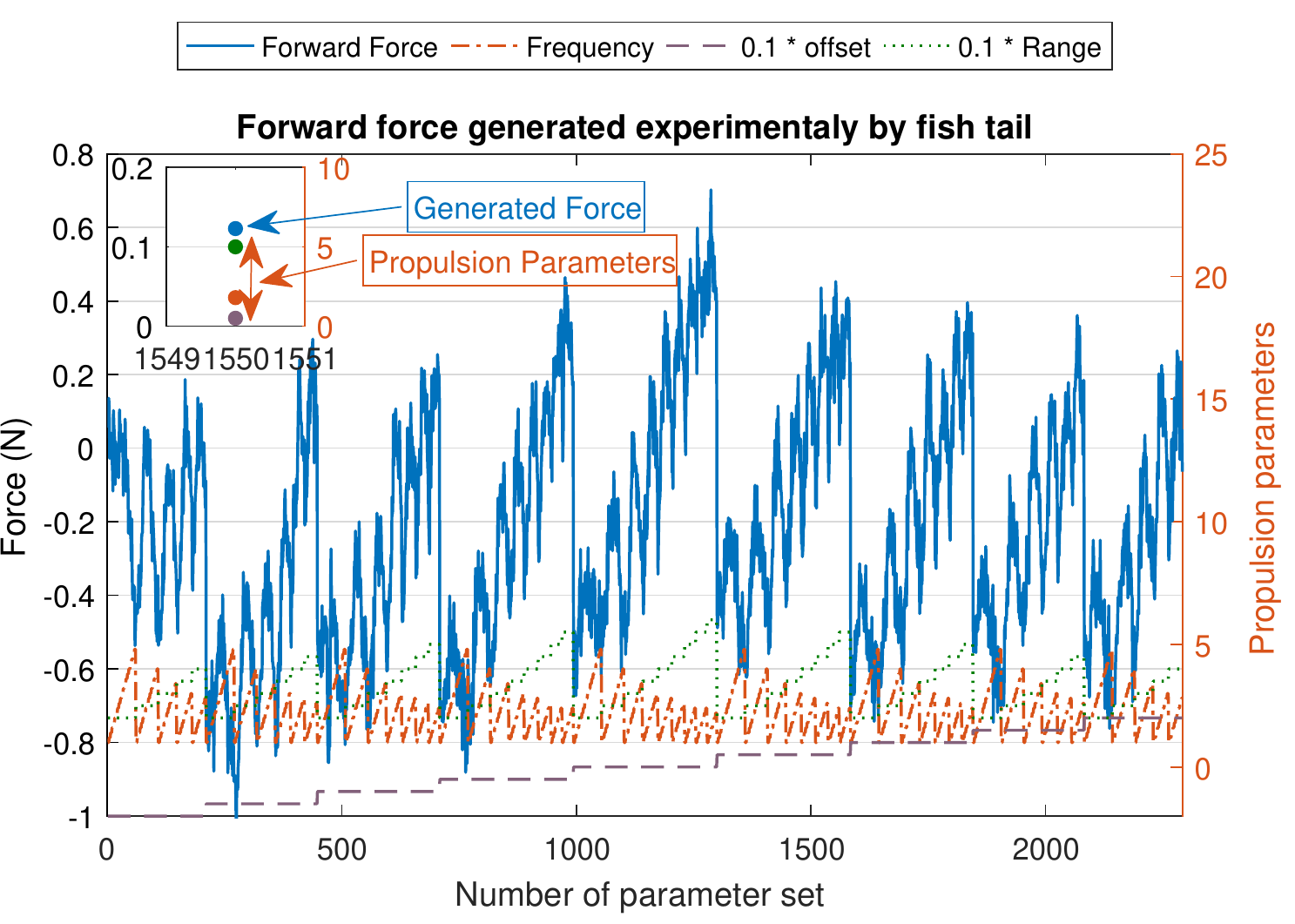}
\label{pic_all_forward}
}
\subfigure[Generated side force as a function of range, offset and frequency]{\includegraphics[scale = 0.6]{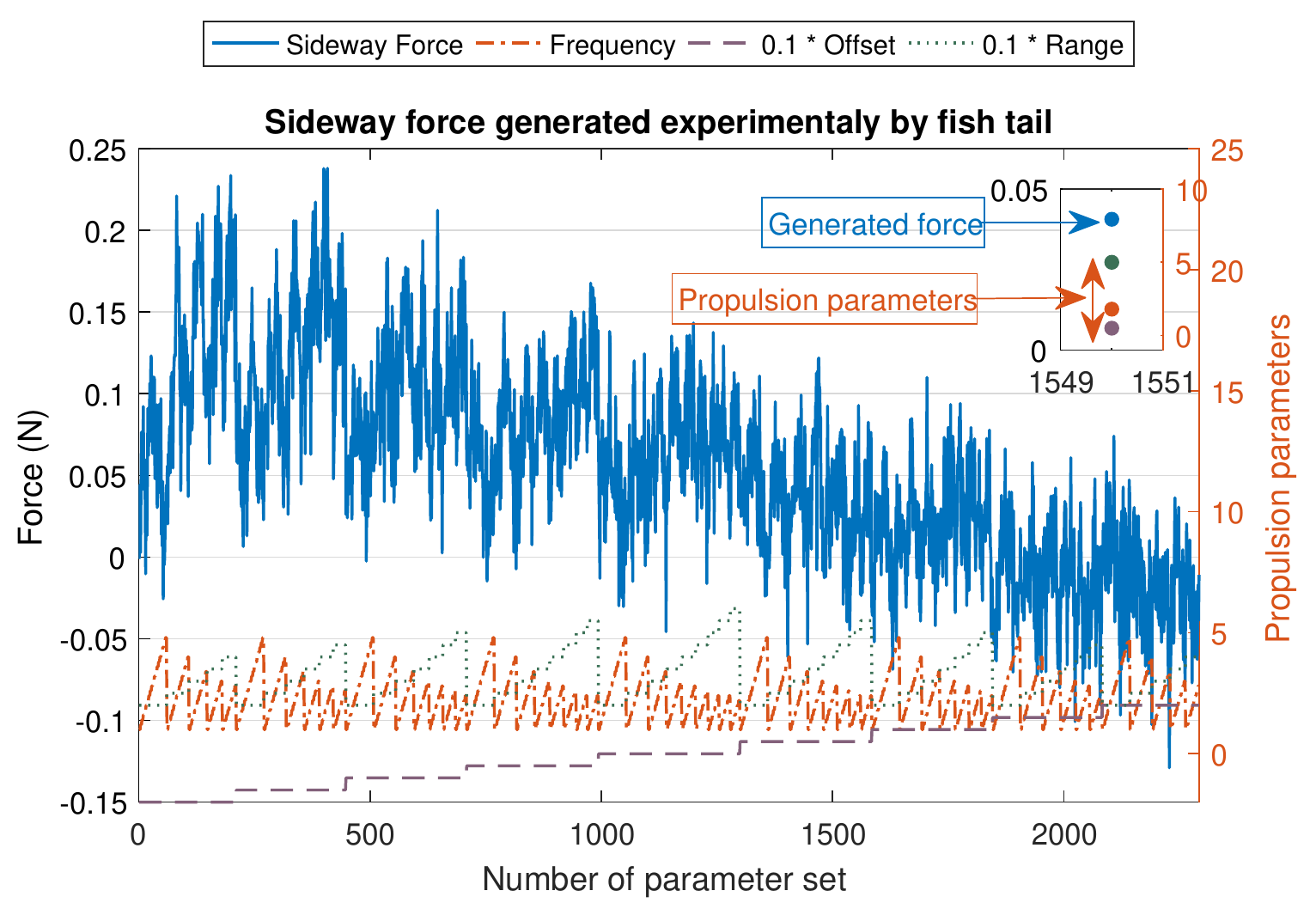}
\label{pic_all_sideway}
}
\caption{Force generated by different fin propulsion.}
%
%
\end{figure*}

In order to identify the system, the 3 dimensional space of $a,b$ and $f$ is spanned (Eq.~\ref{eq_servo}). To this end, for a known set of $a, b$ and $f$, the servomotor is actuated for five cycles and the average forward force and side force are measured. The experiments were performed for $-20^o<b<20^o$, $0<a<±60-abs(b)$ and $0 < f < 82/a$. The limitations on the values of offset, frequency and range are due to limitations in motor maximum speed and the space limitations of the water tank used for experiments. This results in total of 2292 tests. The average force for each test case is presented in Fig.~\ref{pic_all_forward} and Fig.~\ref{pic_all_sideway} with the values of offset $b$, range $a$ and frequency $f$ plotted for reference. As illustrated in small zoomed plot in figures, for each test ($x$ axis data), the servo propulsion parameters is shown(right $y$ axis) as well as the average value of force generated in forward and sideway directions (left $y$-axis).

Based on obtained results, it was found that for the majority of propulsion regimes, the provided force in forward direction is negative. This means that due to special working conditions, in these regimes the propulsion will result in backward movement. This can affect the fish's interaction with its environment. In other words, if the fish is used to push an object in a narrow canal with still water, certain command signals will result in the robot failing to apply positive normal forces to desired objects. From a controls perspective, the system is uncontrollable across much of the three-dimensional control space. As a result, in order to control the robot, a proper subspace with a controllable working regime should be selected and identified. As the goal is to interact with the environment, it is convenient to select a working regime in which the forces which are generated are as high as possible. The authors believe that the reason for negative generated forces lies in the reflected waves coming back from tank walls.

\begin{figure}[t]
\centering
\subfigure[Forward force generated by a symmetric sinusoidal propulsion]{\includegraphics[scale=0.5]{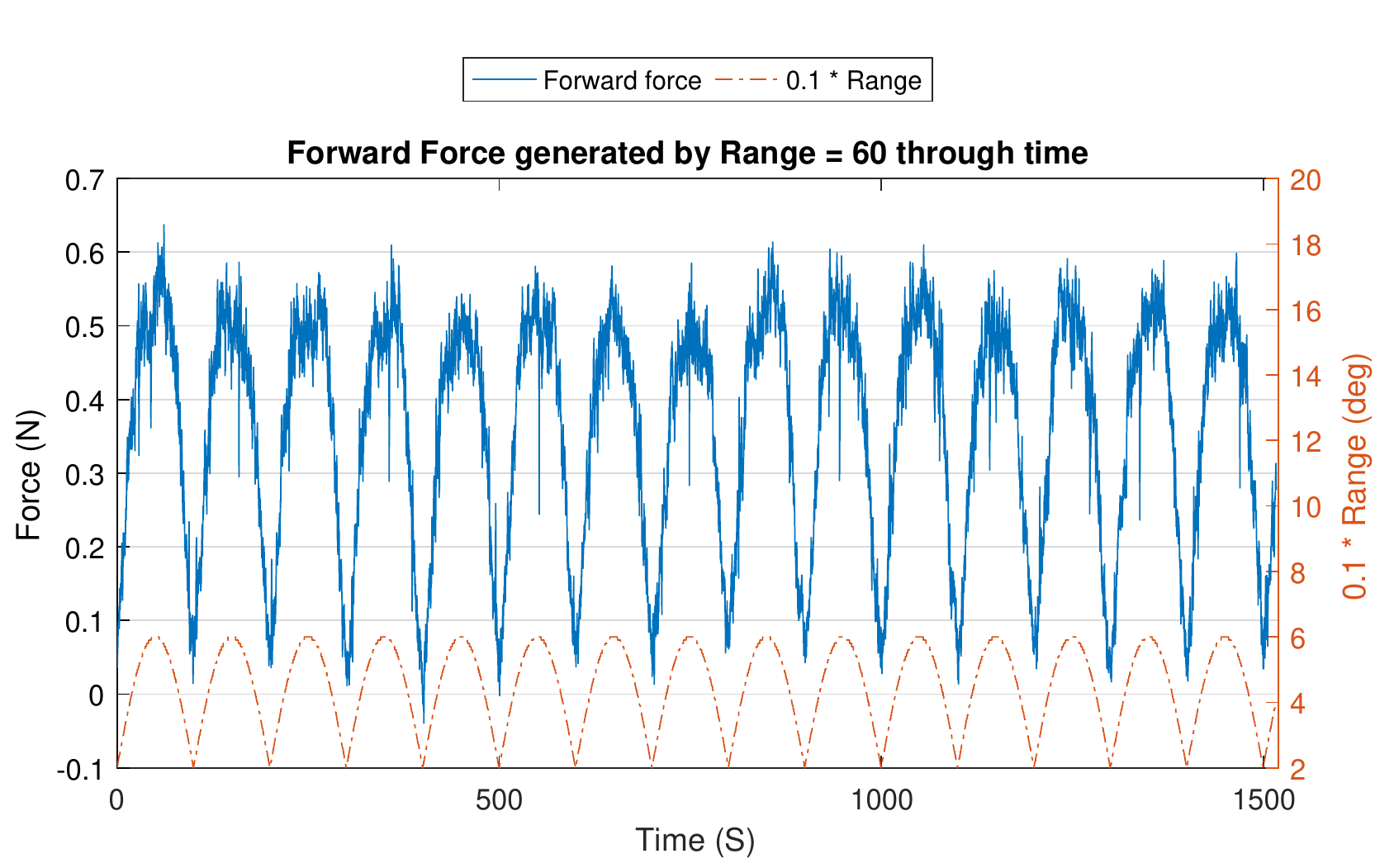}
\label{pic_60_forward}
}
\subfigure[Sideway force generated by a symmetric sinusoidal propulsion]{\includegraphics[scale = 0.5]{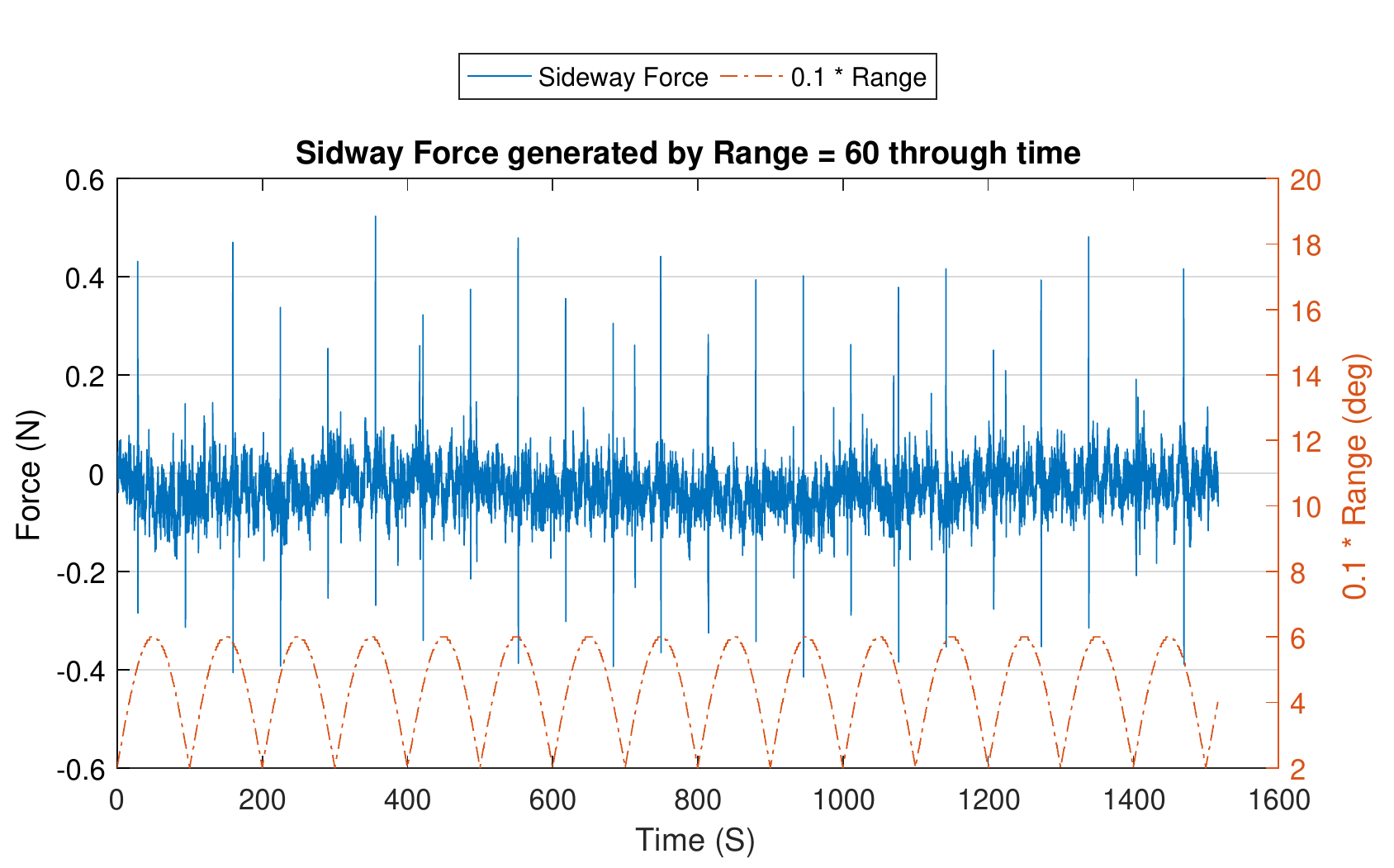}
\label{pic_60_sideway}
}
\caption{Force generated by symmetr ic sinusoidal amplitude fin propulsion. It should be mentioned that $x$axis of plots are not time, but is the number of test, as mentioned in the body of the paper.}
\end{figure}

The maximum forward force is generated with $f =1.4$Hz, $a = 60^\circ$ and $b = 0^\circ$. Searching through the results obtained from spanning the control space, the subspace with fixed propulsion frequency of $1.4$Hz is suitable, while the propulsion amplitude is approximately more than $20^\circ$.  
With the frequency fixed, a second set of experiments was performed to study the dynamic behavior of the output force as a result of other input parameters change. To this end, $b$ was set to 0 and the range was changed according to a half-wave sinusoidal function according to the following equation:
\begin{equation}
a = 20 + \text{abs}(40 \sin(2*pi \times 0.005t))
\end{equation}
The results are presented in Figure~\ref{pic_60_forward} and Figure~\ref{pic_60_sideway} for forward and side forces respectively.

The obtained results show a promising, controllable subspace that includes different propulsion regime which are capable of producing different values of force. It should be mentioned that the results show that the sideway force and as a result the angle of produced force is typically zero for symmetric propulsion. 

\begin{figure}[t]
\centering
\subfigure[Forward force generated by a half-wave sinusoidal range and a full-wave sinusoidal offset input]{\includegraphics[scale=0.42]{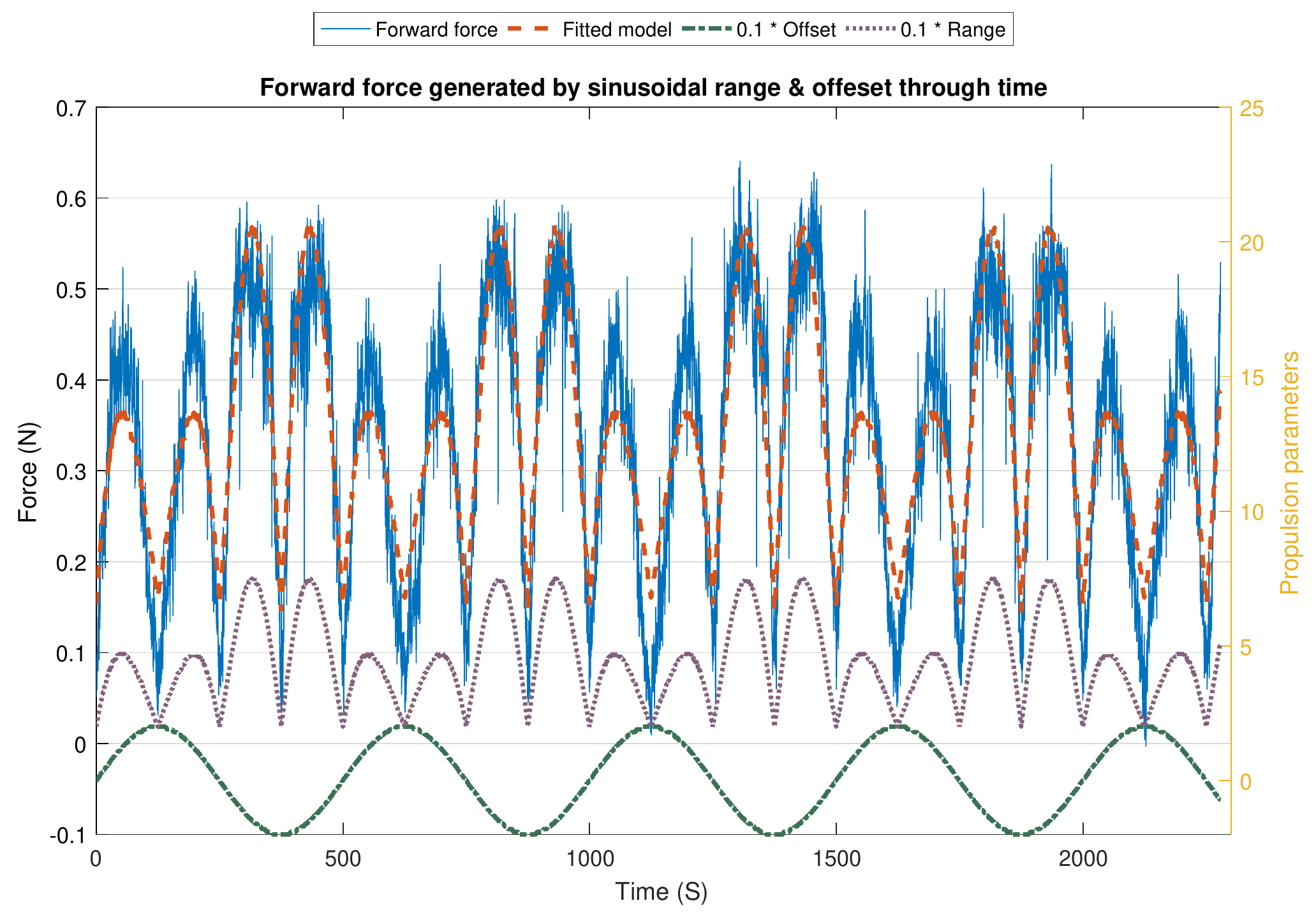}
\label{pic_f_fixed_forward}
}
\subfigure[Side force generated by a half-wave sinusoidal range and a full-wave sinusoidal offset input]{\includegraphics[scale = 0.42]{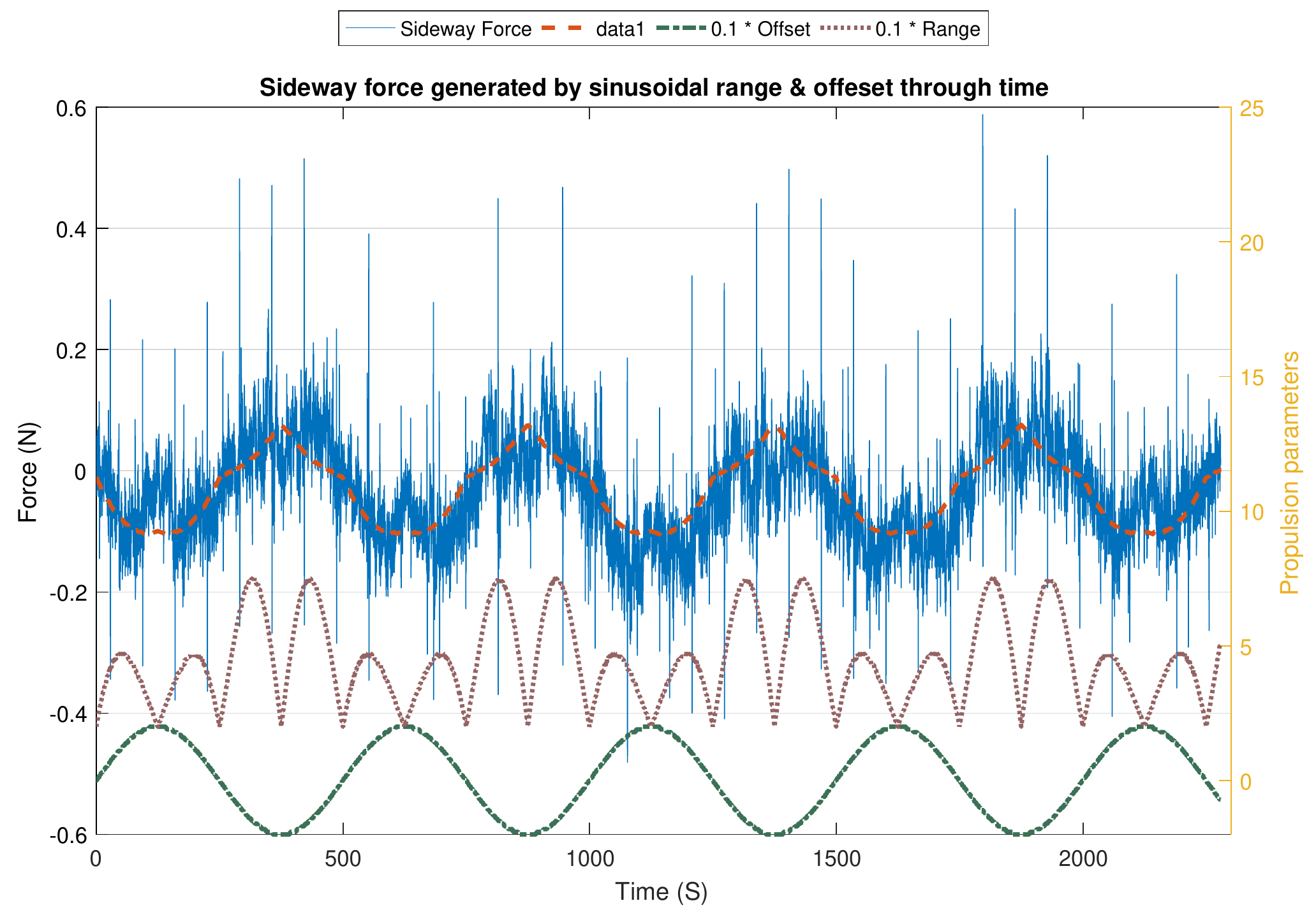}
\label{pic_f_fixed_sideway}
}
\caption{Force generated by sinusoidal offset, sinusoidal amplitude fin propulsion.}
\end{figure}

To further understand the relationship between propulsion parameters and generated force, with frequency set at 1.4 Hz, the propulsion range was changed according a half-wave sinusoidal function, while, simultaneously, its offset was change according a full-wave sinusoidal function:
\begin{equation}
\begin{split}
&\theta = b + a \sin(2\pi \times 1.4t)\\
&b = 20 sin(2*pi \times 0.002t)\\
&a = 20 + \text{abs}((40-b) * sin(2*pi \times 0.004t))
\end{split}
\end{equation}
It should be mentioned that the frequency of change in amplitude and offset (0.002Hz and 0.004Hz, respectively) were selected differently in order to help with distinguishing their effects. The results are presented in Fig.~\ref{pic_f_fixed_forward} and Fig.~\ref{pic_f_fixed_sideway}, showing that the amount of forward force is highly correlated by propulsion range, while the sideway force is mainly effected by the offset of asymmetric propulsion.

In order to model the effect of propulsion parameters on forward and sideway forces, a model was fit to the experimental data using a least-squares approximation plotted in Fig.~\ref{pic_f_fixed_forward} and Fig.~\ref{pic_f_fixed_sideway}. In matrix form, this can be represented as the following\cite{LS}:
%
\begin{equation}
\begin{split}
 \boldmath{Y}  = \boldmath{X} \hat{a}   \quad \Rightarrow \hat{a} ={(\boldmath{X}^T \boldmath{X})}^{-1} \boldmath{X}^T \boldmath{Y}
\end{split}
\end{equation}
\begin{table}
\centering
\caption{Choosing $\boldmath{X}$ basket}
\label{table_basket}
\begin{tabular}{>{\centering\arraybackslash}m{2.5cm} >{\centering\arraybackslash}m{1.5cm} >{\centering\arraybackslash}m{1.5cm}}
\hline
Elements 
& 
FF MAE
&
SF MAE
\\ \hline
$\{a, b\}$
&
7.89\%
&
4.19\%
\\ \hline
$+\{\dot{a}, \dot{b}\}$
&
7.89\%
&
4.19\%
\\ \hline
$+\{\ddot{a}, \ddot{b}\}$
&
7.89\%
&
4.19\%
\\ \hline
$+\{a^2 b^2\}$
&
7.05\%
&
3.98\%
\\ \hline
\color{red}
$+\{a^3 b^3\}$
&
\color{red}4.53\%
&
\color{red}3.83\%
\\ \hline
\color{red}$+\{a^4 b^4\}$
&
\color{red}4.52\%
&
\color{red}3.76\%
\\ \hline
\color{red}$\{a \dot{a} \dot{a}^2 b \dot{b} \dot{b}^2\}$
&
\color{red}7.89\%
&
\color{red}4.19\%
\\ \hline
\end{tabular}
\end{table}

In order to determine what should be in the basket of the fitted model, different combinations were used (Table~\ref{table_basket}). With considering mean absolute error (MAE) values, it can be concluded that a model with third-order terms provides the best MAE, however, the condition number of $(\boldmath{X}^T \boldmath{X}) $matrix containing input basket showed that the resulting matrix is close to a singularity and the data are not reliable. As the addition of second order does not improve the model significantly, the best model for the system is considered to be the first order model of $\{a, b\}$.

The obtained model for forward force along with the original data is plotted in Fig.~\ref{pic_f_fixed_forward} with $a$ and $b$ both changing, while the obtained model for side force along with the original data is plotted in Fig.~\ref{pic_f_fixed_sideway} with $a$ and $b$ both changing.


The identification results can be summarized by saying that in fixed propulsion frequency of $1.4$Hz and propulsion amplitude larger than $20^{\circ}$, the system has a proper behavior with a set of decoupled, linear input-output relations of propulsion amplitude-Force amplitude $\&$ propulsion offset-force angle. These results support the principle of a tail-driven robot to control thrust and direction forces.

\section{Controller Design}
\begin{figure}[t]
\centering
\subfigure[Force magnitude response to a step range input]{\includegraphics[scale=0.5]{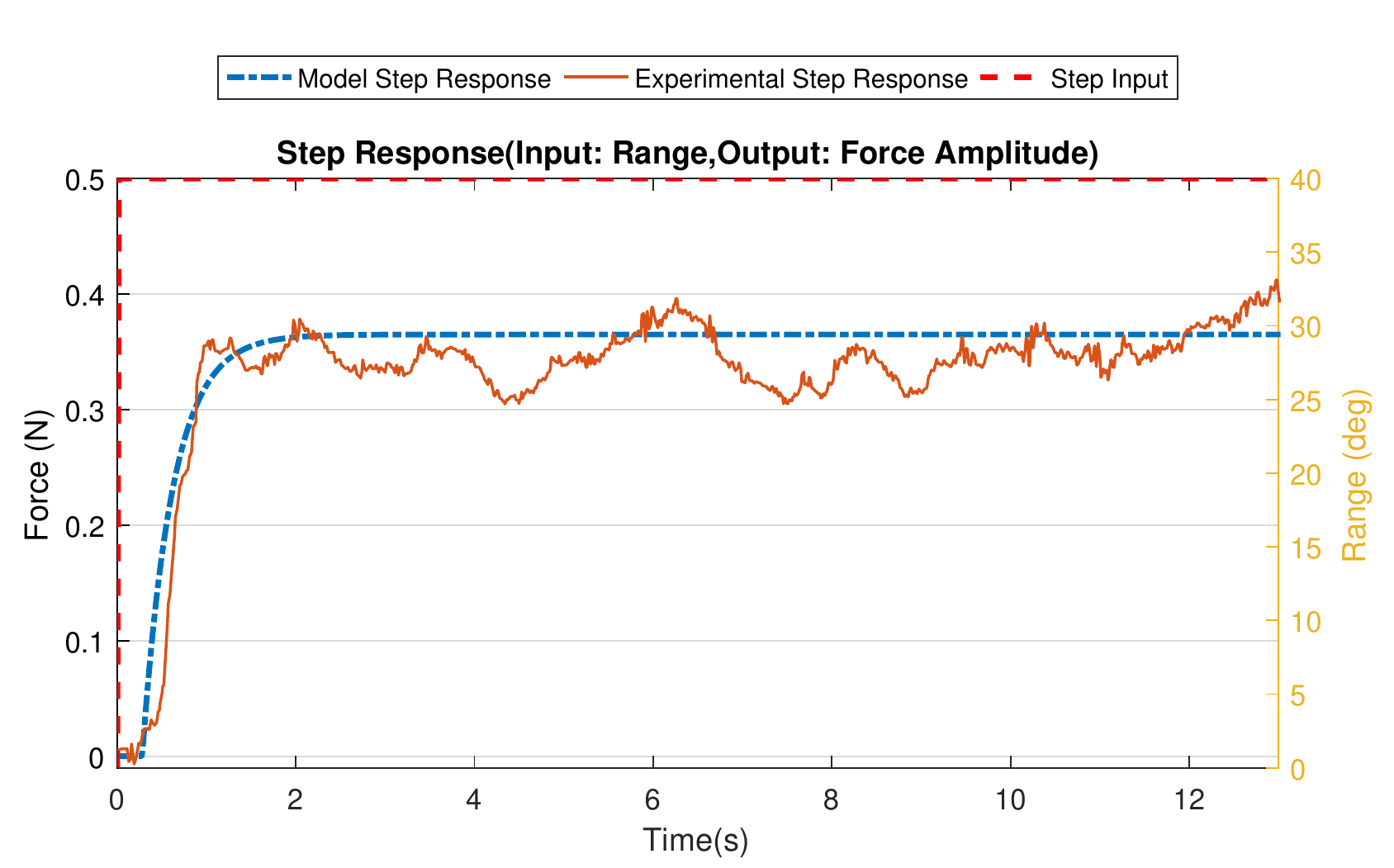}
\label{pic_amp_step}
}
\subfigure[Force angle response to a step offset input]{\includegraphics[scale = 0.5]{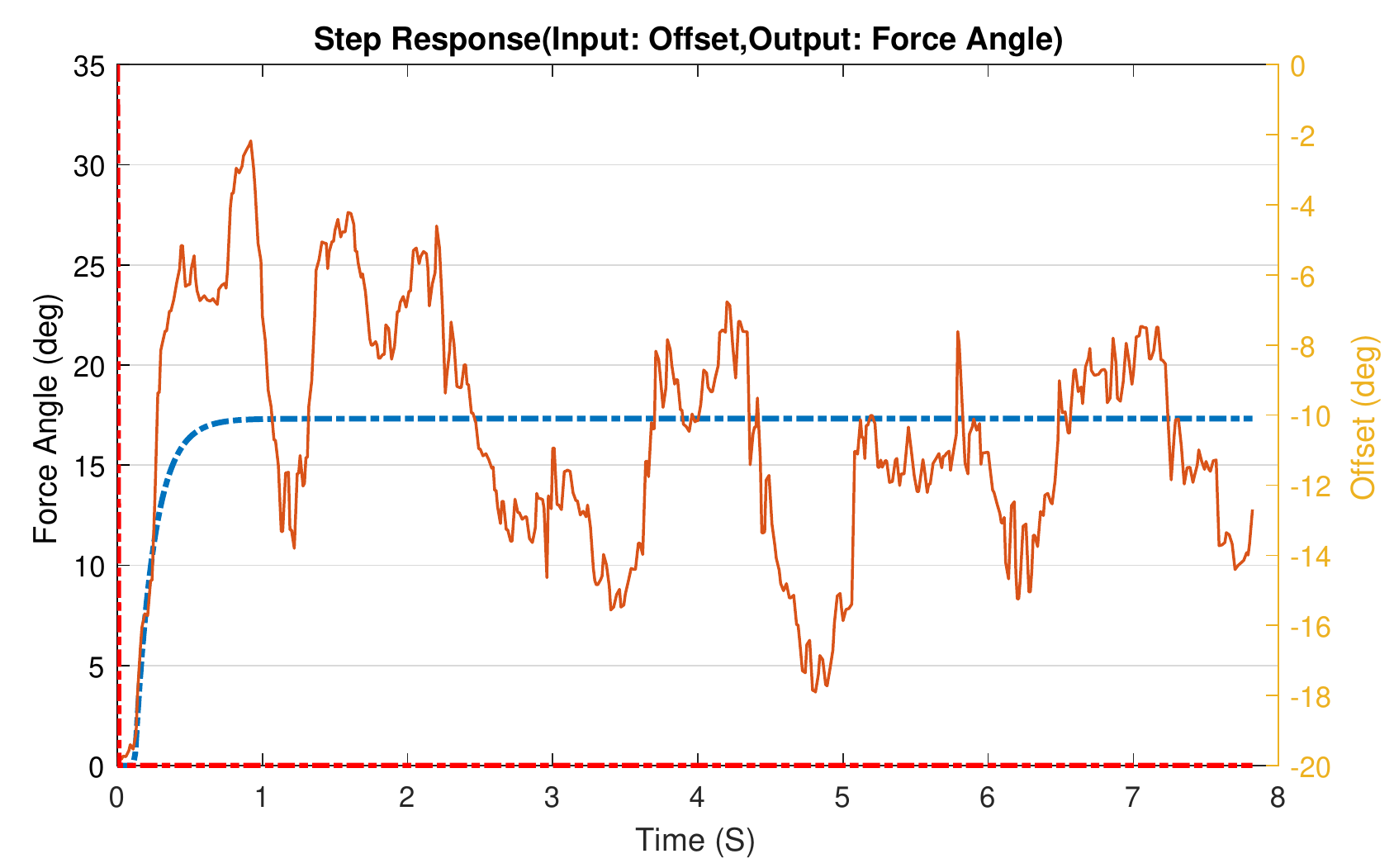}
\label{pic_angle_step}
}
\caption{Step response identification on generated force amplitude \& angle as a function of propulsion amplitude \& offset, respectively}
%
%
\end{figure}

While the previous identification results provide a model for the average forces generated by fin propulsion, a dynamic model is also required for control procedures.  Hence, the system response to a step input was obtained. For this purpose, the force is represented as magnitude and direction values instead of forward and side force. Regarding force magnitude, propulsion amplitude was set to $40^{\circ}$ from stationary and the resulting force magnitude was recorded. In the case of step response identification of force angle, while propulsion amplitude is fixed, the offset was changed from 0 to $-20^{\circ}$ and the resulting force angle was recorded.  Figures~\ref{pic_amp_step}~and~\ref{pic_angle_step} show the step response tests for the force magnitude and angle respectively.
%
%

Based on the obtained results, a first-order transfer function with time delay is fitted to force magnitude step response:
\begin{equation}
%
%
TF_{\text{Forward force}} = \frac{0.00912 e^{-0.28S}}{0.34S+1}
\end{equation}

\begin{figure}[t]
\centering
\subfigure[Closed-loop system block diagram.]{\includegraphics[scale=0.5]{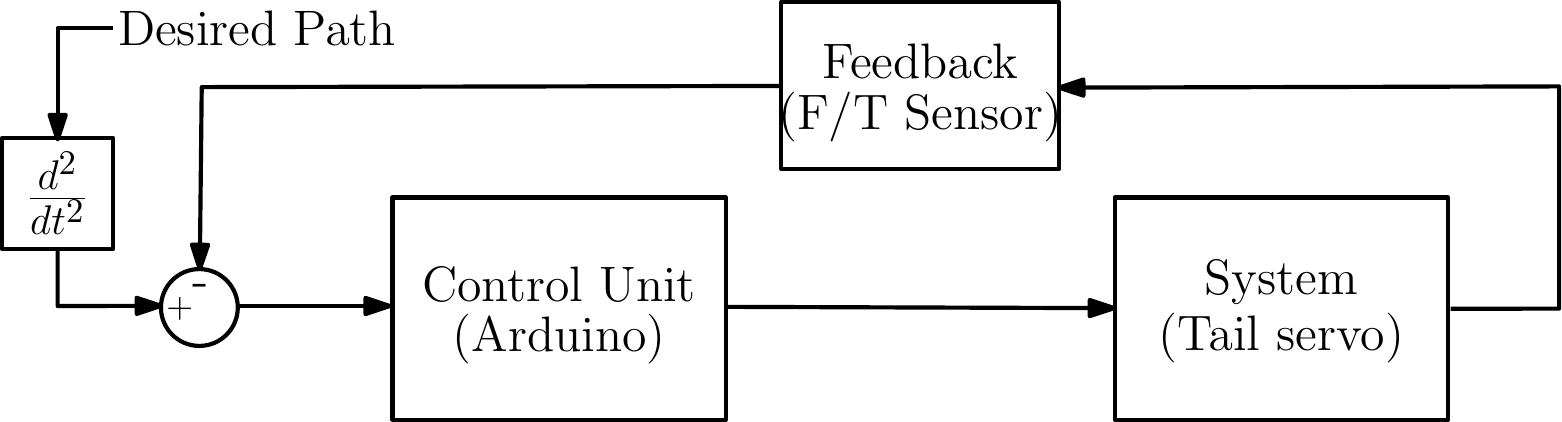}
\label{pic_closed_loop}
}
\subfigure[Block diagram of the system and controller with Feedforward unit.]{\includegraphics[scale = 0.4]{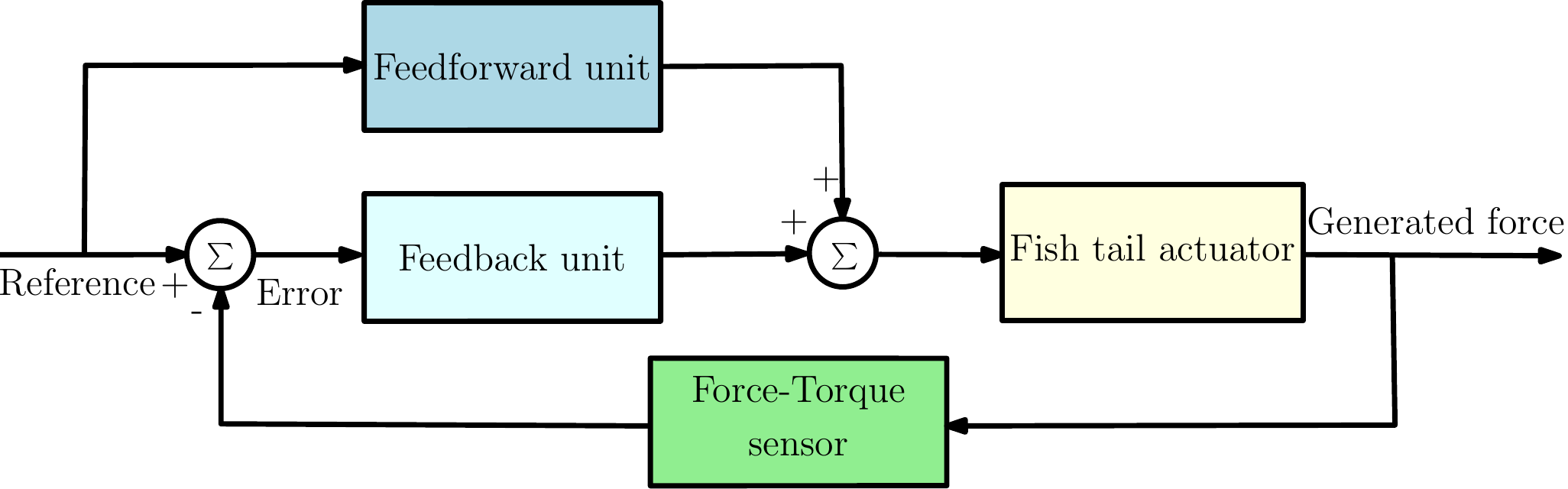}
\label{pic_FF_FB_block}
}
\caption{Block diagram of different controller proposed in controlling force generated by fin propulsion }
\end{figure}

Similarly, a same transfer function is fitted on the force direction:
%
%
\begin{equation}
TF_{\text{Sideway force}} = \frac{-0.866 e^{-0.12S}}{0.13S+1}
\end{equation}
These transfer functions are used in the system block in the block diagram of Fig.~\ref{pic_closed_loop} alongside the system model for tuning control unit.

Evaluation of the results obtained for the  sideway force amplitude (Fig.~\ref{pic_f_fixed_forward}) and force angle (Fig.~\ref{pic_f_fixed_sideway}) show that in the studied working environment, the caudal fin propulsion is not able to provide consistently high forces for sharp turns in a narrow canal. It was also confirmed by a mobile prototype swimming test\footnote{It should be mentioned that applying controllers robust to uncertainties, such as sliding mode controller, was not applicable due to high noise in feedback force}. As a result, the focus of controlling algorithms is the magnitude of force generated by propulsion.

As mentioned before, the identification and control strategy used in this paper is based on the amount and direction of generated force in a full cycle of fin propulsion. As a result, the data received from feedback unit at any time should be averaged over a whole cycle. Hence, a slave parallel process is defined in Python that works alongside of the main Python process and is responsible for determining the average magnitude and direction of the force generated within a single propulsion cycle (Appendix 1).

\begin{figure}[t]
\centering
\subfigure[The input half-sine wave reference force and the generated force of a P controller]{\includegraphics[scale=0.5]{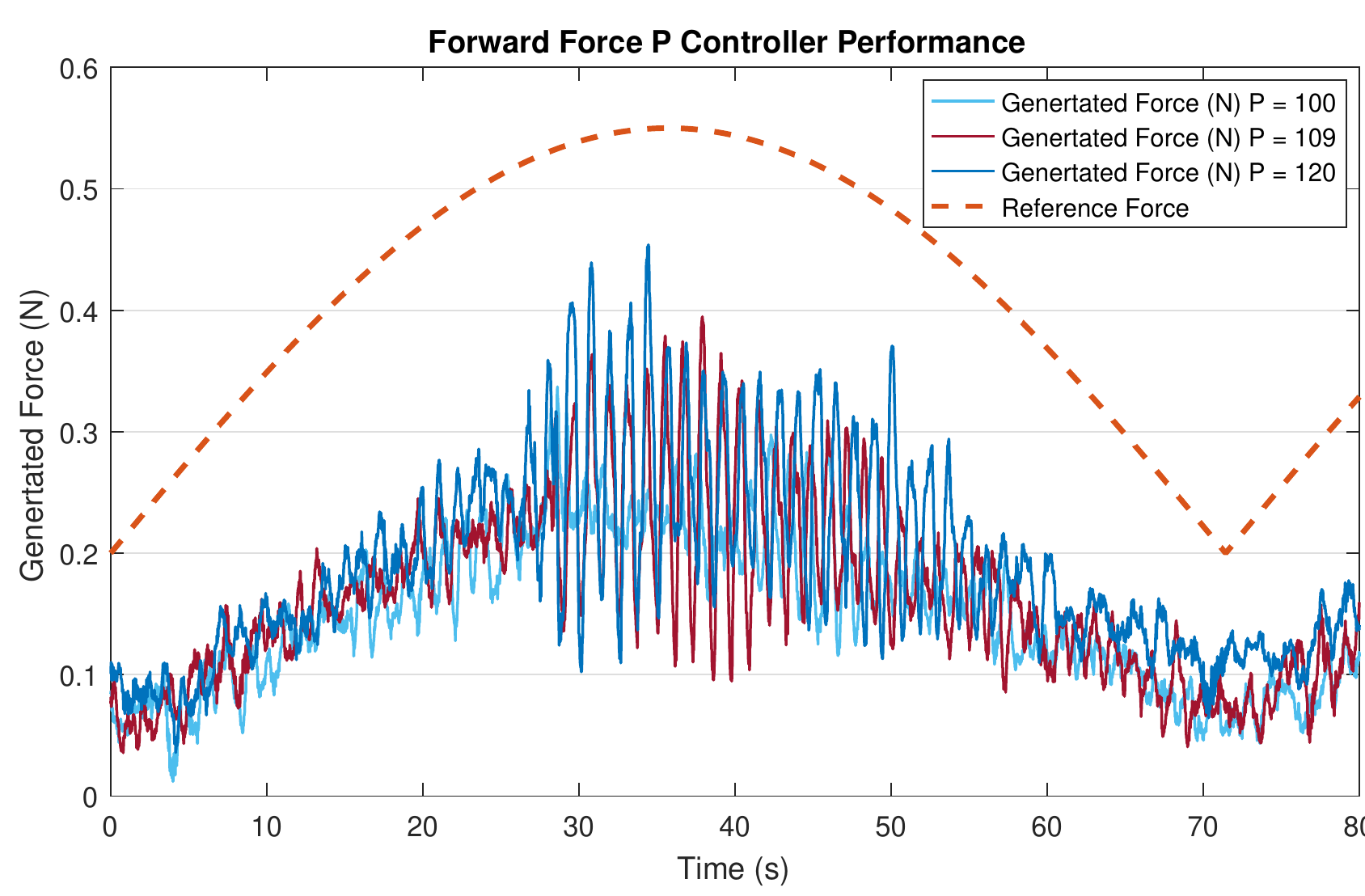}
\label{pic_P}
}
\subfigure[The input half-sine wave reference force and the generated force of a PI controller.]{\includegraphics[scale = 0.5]{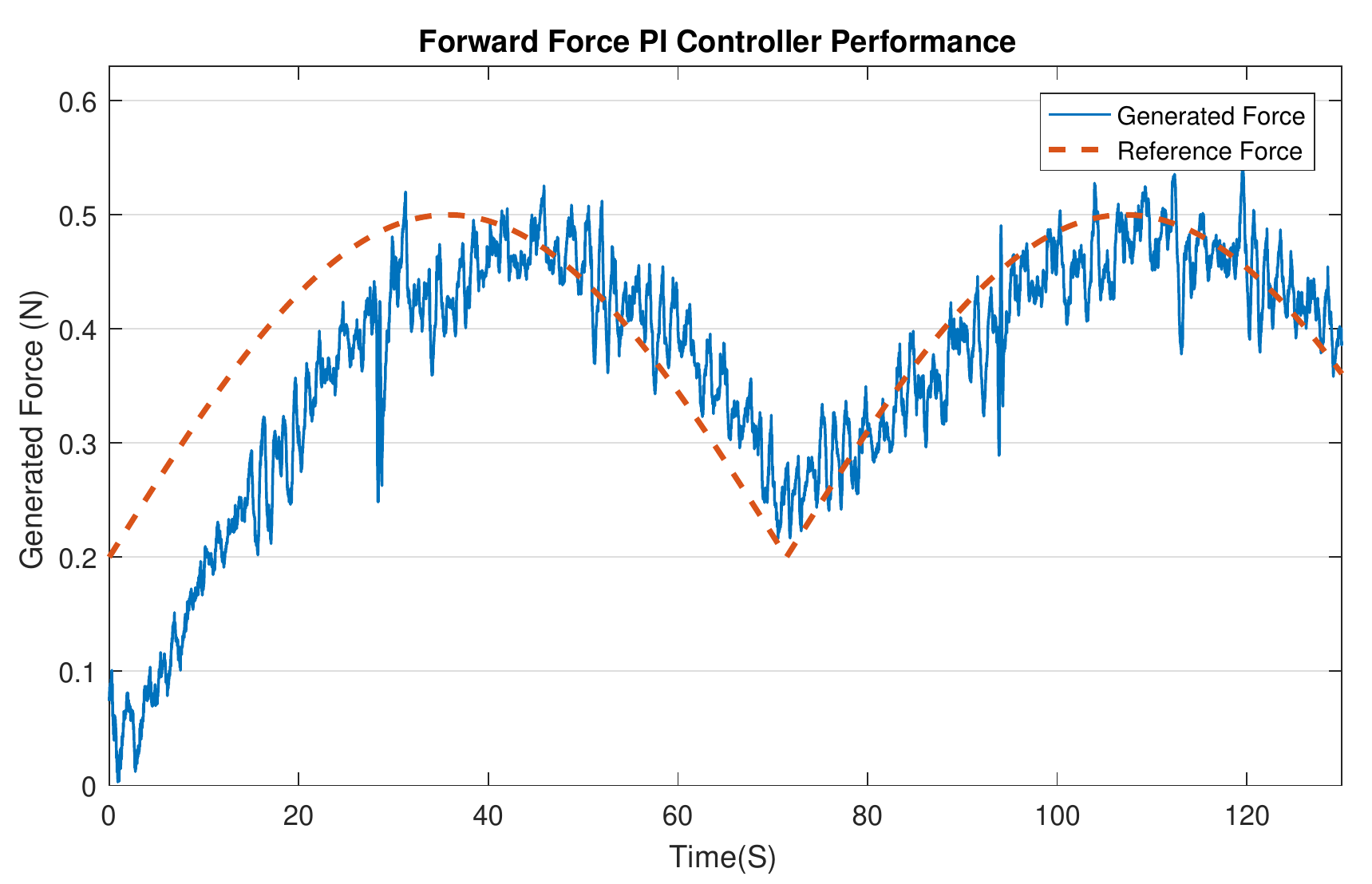}
\label{pic_PI}
}
\caption{PID controller performance.}
\end{figure}

In order to provide a proper controller, the parameters of different PID controllers were tuned using the MATLAB Simulink PID tuning tool in order to be used in control system and then, due to modeling error, tuned more exclusively by experimental tests. A P controller was implemented in the first step. Figure~\ref{pic_P} shows the generated force when the input reference force is a half wave sine signal. As can be seen, the P controller is not able to track the reference input force and have a considerable error for different gains.


To improve the performance, an integrator was added to the controller. However, the experimental results which are shown in Figure~\ref{pic_PI} show that the system is not able to track the input reference force signal in the first cycle. However, after one cycle, the input tracking is able to follow the reference input force. It should be mentioned that due to high noise in force feedback, the addition of a derivative term to the PID controller had a negative effect in controller performance.

\begin{figure}[t]
\centering
\subfigure[The input half-sine wave reference force and the generated force of a combination of a P controller and a feed forward unit.]{\includegraphics[scale=0.5]{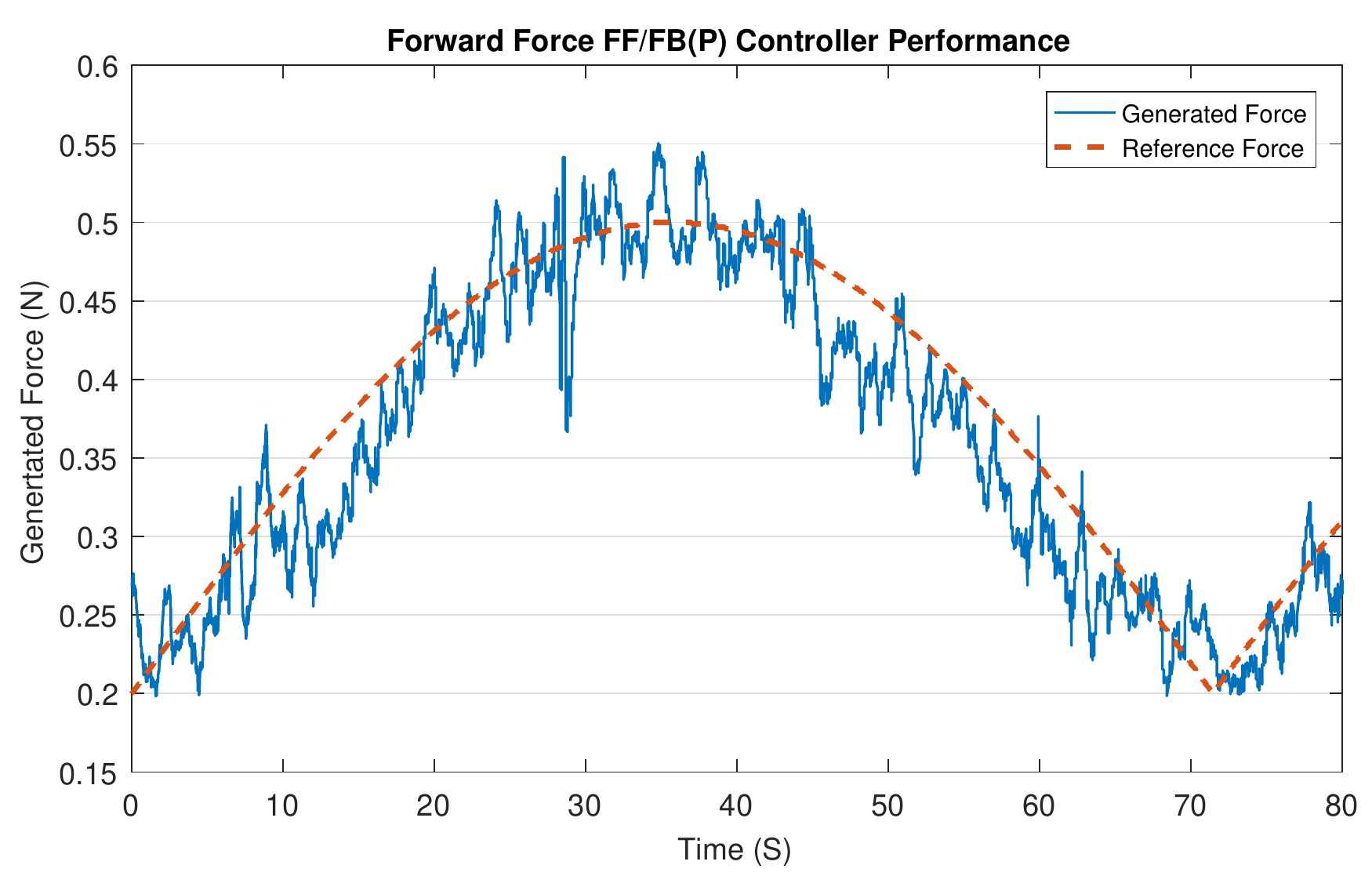}
\label{pic_FF_FB_P}
}
\subfigure[The input half-sine wave reference force and the generated force of a combination of a PI controller and a feed forward unit.]{\includegraphics[scale = 0.5]{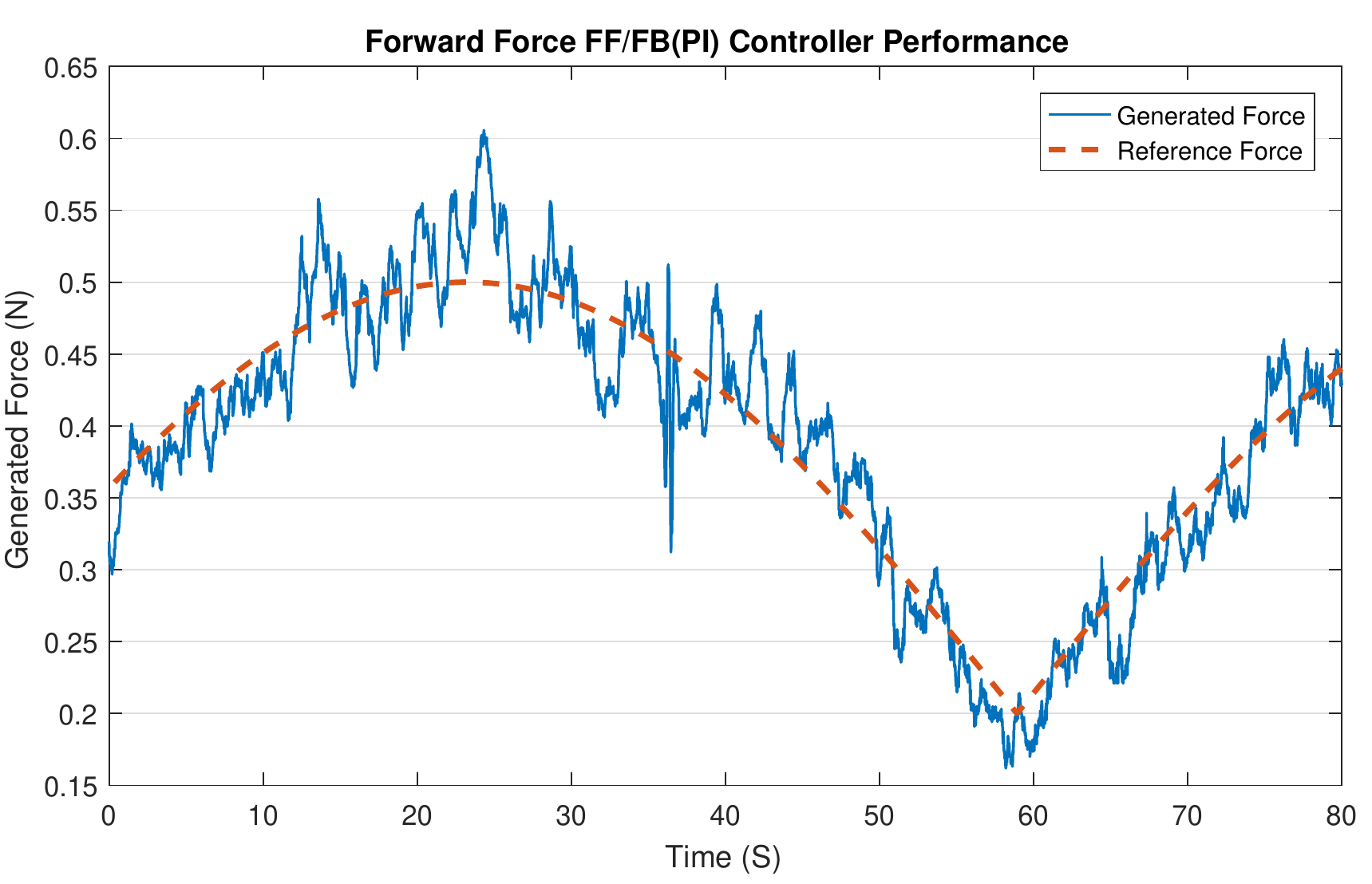}
\label{pic_FF_FB_PI}
}
\caption{Feedforward-feedback controller performance}
%
%
\end{figure}

Further improvements were subsequently achieved by adding a feed forward unit which is shown in Fig.~\ref{pic_FF_FB_block}. To this end, the model obtained from least square identification is used in the feedforward unit.  
As illustrated in Fig.~\ref{pic_FF_FB_P}, when the feedforward unit is added to the system, even a simple P controller is able to provide acceptable performance in tracking desired force. However, it can be seen that during more of the cycle, generated force stays slightly smaller than the desired force. In the controller with both feedforward and PI units, not only is the generated force able  to follow the reference input from the first cycle, but its trend matches the desired force more precisely (Fig.~\ref{pic_FF_FB_PI}).
The PID parameters for controllers with best performances used in the designed controller are summarized in table~\ref{table_PID}. Obtained data shows a noticeable decrement in the PID parameters values when the feedforward unit is added, which shows the positive contribution of this unit.

\begin{table}
\centering
\caption{PID parameters values for different controllers reported in the paper}
%
%
\label{table_PID}
\begin{tabular}{>{\centering\arraybackslash}m{2.5cm} >{\centering\arraybackslash}m{1.5cm} >{\centering\arraybackslash}m{1.5cm} >{\centering\arraybackslash}m{1.5cm}}
\hline
Controller 
& 
P value
&
I value
&
D value
\\ \hline
P
&
100,109,120
&
0
&
0
\\ \hline
PI
&
80
&
0.12
&
0
\\ \hline
FF-P
&
10.9
&
0
&
0
\\ \hline
FF-PI
&
9
&
0.05
&
0
\\ \hline
\end{tabular}
\end{table}

\section{Conclusion \& Future work}


There are many mechanisms for underwater propulsion. Bio-inspired methods are usually more energy efficient, which can be an important consideration in underwater autonomous vehicles. Based on this assumption, a fish inspired robot was designed and built to work in a narrow and shallow workspace such as a small canal. 

Our main goal in this research was to find the best control strategy for the navigation of the fish in the canal. For this purpose, a test set up was designed and used for data acquisition and control. Since there are many uncertainties with underwater robotic fish working in narrow canals due to turbulence and hydrodynamic effects, building an accurate analytic model of the system is highly complicated. The methods introduced in this paper have been used to identify the relationships between the amplitude and frequency of the tail propulsion and magnitude and direction of the forces generated. 

Experimental results showed that the model obtained for the generated thrust was linear with a good approximation when the propulsion magnitude varies linearly within the controllable regime.  We can conclude that the caudal fin of the current generation of robot is not capable of producing enough consistent sideway force for sharp turns in narrow working environments.

Several linear controllers were tested to find out the best solution for generating directed thrust. The data shows that the PI controller with a feed forward unit is the best strategy to track the reference input force with acceptable accuracy. The obtained results demonstrate that by applying the feedforward feedback controller, it is possible to produce desired amounts of forward force by caudal fin propulsion even in narrow environments.

Future work includes using a higher-resolution, lower-noise force sensor which may permit selecting different control algorithms such as a sliding-mode controller to overcome the uncertainties in force angle. Moreover, a pair of pectoral fins will to be added to the fish in the next prototype in order to enable sharp turns.  
Biological studies have shown that the pectoral fins are also useful for generating thrust and are even used as main locomotion for low speed movement in fishes. This addition will permit our team to study simultaneous pectoral and caudal fin propulsion.

\begin{acknowledgment}
This work was supported in part by Salt River Project.
\end{acknowledgment}
\newpage

\appendix       
\section*{Appendix A: Data Flow In control system}
\begin{algorithm}[h]
\caption{\textbf {Data Flow Between Control Unit, FT sensor \& Arduino}}
\label{code_Data}
\begin{algorithmic}[1]
\Statex \begin{center} \textbf{Data Communication Master Process}\end{center}
\State Initialize Arduino. \Comment{Connect}
\State Initialize Force-Torque sensor. \Comment{Connect, begin stream \& calibrate}
\State Save calibration values.
\Procedure{Main Parallel Process}{Actuation \& Data Sampling}
\State Set the frequency of tail flapping.
\State Define \& begin slave process as a parallel process.
\State Start timer
\While{$T<T_f$} \Comment{$T_f$ is final time}
\State Command Arduino to actuate the tail servo.
\State Ask the slave process for the most recent data.
\State Log \& save data.
\EndWhile
\State Terminate the slave process.
\State Disconnect from Arduino.
\State Disconnect from Force-Torque sensor.
\EndProcedure
\Procedure{Slave Parallel Process}{FT Sensor Data processing}
\State Get the frequency of tail flapping.
\State Get the calibration values.
\State Calculate the number of data in each flapping cycle ($n_{cycle}$). \Comment{Based on the streaming frequency}
\State Get $n_{cycle}$ number of data and save them in a matrix. \Comment{By considering the calibration values}
\While{1}
\State Update the vector with the newest data.
\State Calculate the mean of each matrix vectors.
\State Save the mean values so it reachable for main process.
\EndWhile
\EndProcedure
\end{algorithmic}
\end{algorithm}





\newpage

\bibliographystyle{asmems4}

\bibliography{mybib}

\end{document}